# Joint Coreference Resolution for Zeros and non-Zeros in Arabic

Abdulrahman Aloraini[1,2]   Sameer Pradhan[3,4]   Massimo Poesio[1]

[1]School of Electronic Engineering and Computer Science, Queen Mary University of London
[2]Department of Information Technology, College of Computer, Qassim University
[3]cemantix.org
[4]Linguistic Data Consortium, University of Pennsylvania, Philadelphia, USA

a.aloraini@qmul.ac.uk   spradhan@$\frac{\text{cemantix.org}}{\text{upenn.edu}}$   m.poesio@qmul.ac.uk

## Abstract

Most existing proposals about anaphoric zero pronoun (AZP) resolution regard full mention coreference and AZP resolution as two independent tasks, even though the two tasks are clearly related. The main issues that need tackling to develop a joint model for zero and non-zero mentions are the difference between the two types of arguments (zero pronouns, being null, provide no nominal information) and the lack of annotated datasets of a suitable size in which both types of arguments are annotated for languages other than Chinese and Japanese. In this paper, we introduce two architectures for jointly resolving AZPs and non-AZPs, and evaluate them on Arabic, a language for which, as far as we know, there has been no prior work on joint resolution. Doing this also required creating a new version of the Arabic subset of the standard coreference resolution dataset used for the CoNLL-2012 shared task (Pradhan et al., 2012) in which both zeros and non-zeros are included in a single dataset.

## 1 Introduction

In pronoun-dropping (pro-drop) languages such as Arabic (Eid, 1983), Chinese (Li and Thompson, 1979), Italian (Di Eugenio, 1990) and other romance languages (e.g., Portuguese, Spanish), Japanese (Kameyama, 1985), and others (Young-Joo, 2000), arguments in syntactic positions in which a pronoun is used in English can be omitted. Such arguments–sometimes called null arguments, empty arguments, or zeros, and called anaphoric zero pronouns (AZP) here when they are anaphoric, are illustrated by the following example:

... المفارقة الأخرى عن بوش هي عدم حماسته للمؤتمر الدولي ، لأن بوش من البداية ، يريد * اجتماعا مختلفا ...

*Ironically, Bush did not show any enthusiasm for the international conference, because Bush since the beginning, (he) wanted to attend another conference ...*

In the example, the '*' is an anaphoric zero pronoun–a gap replacing an omitted pronoun which refers to a previously mentioned noun, i.e. Bush.[1]

Although AZPs are common in pro-drop languages (Chen and Ng, 2016), they are typically not considered in standard coreference resolution architectures. Existing coreference resolution systems for Arabic would cluster the overt mentions of Bush, but not the AZP position; vice versa, AZP resolution systems would resolve the AZP, to one of the previous mentions, but not other mentions. The main reason for this is that AZPs are empty mentions, meaning that it is not possible to encode features commonly used in coreference systems–the head, syntactic and lexical features as in pre-neural systems. As a result, papers such as (Iida et al., 2015) have shown that treating the resolution of AZPs and realized mentions separately is beneficial. However, it has been shown that the more recent language models and end-to-end systems do not suffer from these issues to the same extent. BERT, for example, learns surface, semantic and syntactic features of the whole context (Jawahar et al., 2019) and it has been shown that BERT encodes sufficient information about AZPs within its layers to achieve reasonable performance (Aloraini and Poesio, 2020b,a). However, these findings have not yet led to many coreference resolution models attempting to resolve both types of mentions in a single learning framework (in fact, we are only aware of two, (Chen et al., 2021; Yang et al., 2022), the second of which was just proposed) and these have not been evaluated with Arabic.

In this paper, we discuss two methods for jointly clustering AZPs and non-AZPs, that we evaluate on Arabic: a *pipeline* and a *joint learning* architecture. In order to train and test these two architectures, however, it was also necessary to create a new version of the Arabic portion of the CoNLL-2012 shared task corpus in which both zeros and non-zeros are annotated in the same documents. To summarize, the contributions of this paper are as follows:

- We introduce two new architectures for resolving AZPs and non-AZPs together, the *pipeline* and the *joint learning* architecture. One of our architectures, the *joint learning*, outperforms the one existing joint end-to-end model (Chen et al., 2021) when resolving both types of mentions together.

- We create an extended version of the Arabic portion of CoNLL-2012 shared task in which the zero and non-zero mentions are represented in the same document. The extended dataset is suitable for training AZPs and non-AZPs jointly or each type separately.

---

[1]We use here the notation for AZPs used in the Arabic portion of OntoNotes 5.0, in which AZPs are denoted as **\*** and we also use another notation which is **\*pro\***.

## 2 Related Work

Most existing works regard coreference resolution and AZP resolution as two independent tasks. Many studies were dedicated to Arabic coreference resolution using CoNLL-2012 dataset (li, 2012; Zhekova and Kübler, 2010; Björkelund and Nugues, 2011; Stamborg et al., 2012; Uryupina et al., 2012; Fernandes et al., 2014; Björkelund and Kuhn, 2014; Aloraini et al., 2020; Min, 2021), but AZPs were excluded from the dataset so no work considered them. Aloraini and Poesio (2020b) proposed a BERT-base approach to resolve AZPs to their true antecedent, but they did not resolve other mentions.

There have been a few proposals on solving the two tasks jointly for other languages. Iida and Poesio (2011) integrated the AZP resolver with a coreference resolution system using an integer-linear-programming model. Kong and Ng (2013) employed AZPs to improve the coreference resolution of non-AZPs using a syntactic parser. Shibata and Kurohashi (2018) proposed an entity-based joint coreference resolution and predicate argument structure analysis for Japanese. However, these works relied on language-specific features and some assumed the presence of AZPs.

There are two end-to-end neural proposals about learning AZPs and non-AZPs together. The first proposal is by Chen et al. (2021) who combined tokens and AZP gaps representations using an encoder. The two representations interact in a two-stage mechanism to learn their coreference information, as shown in Figure 5. The second proposal, just published, is by (Yang et al., 2022), who proposed the CorefDPR architecture. CorefDPR consists of four components: the input representation layer, coreference resolution layer, pronoun recovery layer and general CRF layer. In our experiments, we only compared our results with the first proposal because the second system was only evaluated on the Chinese conversational speech of OntoNotes[2] and the model is not publicly available which makes it difficult to compare our results with theirs.

## 3 An Extended Version of the CoNLL Arabic dataset with AZPs

The goal of the CoNLL-2012 coreference shared task is to learn coreference resolution for three languages (English, Chinese and Arabic). However, AZPs were excluded from the task even though they are annotated in OntoNotes Arabic and Chinese. This was because considering AZPs decreased the overall performance on Arabic and Chinese(Pradhan et al., 2012), but not on English because it is not a pro-drop language (White, 1985). So in order to study joint coreference resolution for explicit mentions and zero anaphors, we had to create a novel version of the CoNLL-2012 dataset in which AZPs and all related information are included. The CoNLL-2012 annotation layers consists of 13 layers and they are in Appendix A.

Existing proposals evaluated their AZP systems using OntoNotes Normal Forms (ONF)[3]. They are annotated with AZPs and other mentions; however, they are not as well-prepared as CoNLL-2012. To create a CoNLL-like dataset with AZPs, we extract AZPs from ONF and add them to the already-existing CoNLL files. The goal of the new dataset is to be suitable for clustering AZPs and non-AZPs, and can be compared with previous proposals that did not consider AZPs and as well as with future works that consider them.

To include AZPs and their information (e.g., Part-of-Speech and parse tree) to CoNLL-2012, we can use ONF. However, while adding AZPs to the clusters, we realized that there is one difficulty:some

---

[2]The TC part of the Chinese portion in OntoNotes.
[3]The OntoNotes Normal Form (ONF) was originally meant to be an human-readable integrated representation of the multiple layers in OntoNotes. However, it has been used by many as a machine readable representation–as it is also more or less true–to extract annotations, primarily zeros that are typically excluded from the traditional CoNLL tabular representation.

```
Chain 71 (IDENT)
        6.2-13       الـجَيْشِ الـشَعْبِيّ لِـ -تَحْرِيرِ الـسُودانَ " سمسون خـواجة
        7.2-2        *

Chain 92 (IDENT)
        8.1-11       وِزارَةِ الـخارجِيَّةِ الـسُودانِـيَّةِ مُطْرف صِدِّيق الَّذِي يَرْأَسُ
                     الـحُكُومِيَّ
        8.16-16      ه

Chain 95 (APPOS)
  ATTRIB 8.1-4       وَكِيلُ وِزارَةِ الـخارِجِيَّةِ الـسُودانِـيَّةِ
  HEAD   8.5-6       مُطْرف صِدِّيق
```

Figure 1: A screenshot of OntoNotes Normal Forms (onf). Chain 71 is not considered part of a CoNLL-2012 shared task because the cluster would become a singleton when we remove the AZP (denoted as *).

coreference chains only exist in ONF, but not in CoNLL-2012. These are clusters consisting of only one mention and one AZP, as in the example illustrated in Figure 1. Chain 71 has two mentions, an AZP (denoted with *) and a mention. Since CoNLL-2012 does not consider AZPs in coreference chains, this cluster would only have a single mention because CoNLL-2012 removed AZPs (these clusters are known as singletons, contains only one mention). Our new dataset includes AZPs; therefore, such clusters should be included. To add them to the existing CoNLL-2012, we have to assign them a new cluster. We did this by writing a script that automatically extracts AZPs from ONF and adds them in CoNLL-2012 following these steps:

1. Finds all clusters that have AZPs in ONF and extracts AZPs.

2. Each extracted AZP is either:
   (a) Clustered with two or more mentions: For this case, CoNLL has already assigned a coreference-chain number and we assign the AZP to the same number.
   (b) Clustered with only one mention: We create a new cluster that include the single mention and the AZP.

3. Adds the AZP and writes other relevant information, such as, Part-of-Speech, syntax, and all the annotation layers.

Adding AZPs to CoNL-2012 is beneficial to learn how to resolve them with other mentions or can be useful for future CoNLL-shared tasks and any other related NLP task. After preparing the new CoNLL dataset as discussed, we used it to train the joint coreference model. This new version of Arabic OntoNotes will be made available with the next release of OntoNotes. The distribution of documents, sentences, words, and AZPs of this extended dataset are in Table 1.

| Category  | Training | Development | Test   |
|-----------|----------|-------------|--------|
| Documents | 359      | 44          | 44     |
| Sentences | 7,422    | 950         | 1,003  |
| Words     | 264,589  | 30,942      | 30,935 |
| AZPs      | 3,495    | 474         | 412    |

Table 1: The documents, sentences, words and AZPs of the extended version of CoNLL-2012. We follow the same split as in the original CoNLL-2012 for training, development and test.

## 4 The Models

Earlier proposals resolved AZPs based on the antecedents that are in the same sentence as the AZP or two sentences away (Chen and Ng, 2015, 2016; Yin et al., 2016, 2017; Liu et al., 2017; Yin et al., 2018; Aloraini and Poesio, 2020b). However, it has been shown that learning mention coreference in the whole document is beneficial for AZP resolution (Chen et al., 2021). Therefore, we apply two novel methods for resolving AZPs using clusters and coreference chains. The *pipeline* resolves AZPs based on the output clusters from the coreference resolution model while the *joint learning* learns how to resolve AZPs from the coreference chains, we show an example of these two in Figure 2. In the example, the *pipeline* resolves AZPs to clusters, instead of mentions and the *joint learning* finds the coreference chains for mentions, including AZPs. Earlier proposals suffered from two main problems. First, they consider a limited number of candidates (i.e mentions in two sentences away from the AZP) as possible true antecedents; however, the true antecedent might be far away from the AZP. Second, other mentions can share salient context as the true antecedent which can introduce more noise to the learning. Our methods mitigate these problems by considering all mentions in the document and employing more relevant information. The *pipeline* resolves AZPs based on clusters which decreases dramatically the number of AZP candidates. The *joint learning* resolves AZPs using coreference chains which incorporates broader context for AZPs, insufficient contexts results in many errors (Chen and Ng, 2016).

### 4.1 The Pipeline Model

In a *pipeline* setting, the inputs are the extended version of CoNLL, the one we described in Section 3. Each file consists of multiple sentences and we follow the same splits in CoNLL-2012 (Pradhan et al., 2012) for train, development and test. We initially fed the documents for training into two models: *coreference resolution* and *AZP identification*. We used the Arabic coreference resolution by (Aloraini et al., 2020) and the proposed AZP identification by (Aloraini and Poesio, 2020a). The outputs of coreference resolutions are clusters and each one has its own mentions. The outputs of the AZP identification are the predicted gap positions of AZPs. The *AZP resolution* model by (Aloraini and Poesio, 2020b) learns how to resolve the identified AZPs with their clusters. We show how we represent the input in the following:

The *input* is a document with sentences separated with periods, and has a total of *n* words. The *input* does not consider AZPs initially, they are masked.

$$input = (w_1, w_2, w_3, ..., w_n) \quad (1)$$

We first feed the *input* into the *coreference resolution* model which outputs the mention clusters, $c_1, c_2$, to the last cluster index, $k$.

$$output\_clusters = coref\_res(input) \quad (2)$$
$$output\_clusters = (c_1, c_2, ..., c_k) \quad (3)$$

After finding the coreference clusters, the *AZP Identification* model predicts the AZP positions in two steps. First, the *AZP identification* uses a Part-of-Speech tool to tag words and mark gaps after verbs as potential AZPs. Second, *AZP identification* classifies these marked gaps as AZPs or not.

Therefore, not every gap between words has an AZP. For example, in (5) there is no AZP between the words $w_2$ and $w_3$, but there is one between $w_1$ and $w_2$ (i.e. $a_i$). We find AZP locations and extract their positions.

$$input\_with\_azp = AZP\_Id(input) \quad (4)$$
$$input\_with\_azp = (w_1, a_i, w_2, w_3, ..., w_n) \quad (5)$$
$$AZPs = (a_i, ..., a_k) \quad (6)$$
$$ss = same\_sentence(a_i, c_j) \quad (7)$$
$$cd = azp\_cluster\_distance(a_i, c_j) \quad (8)$$
$$AZP_i = (a_i\_pre, a_i\_next, ss, cd) \quad (9)$$

We follow the same representation for AZPs as (Aloraini and Poesio, 2020b):

- embeddings for previous word to AZP.
- embeddings for next word to AZP.
- Whether the AZP and the candidate entity (represented either as the last mention or first mention) are in the same sentence or not.
- The distance between the AZP and its cluster representation.

The four features are concatenated, as shown in (9).

Clusters can be represented in different ways, including, e.g, the representation of the first mention or the last mention. We found empirically that representing clusters with the nearest mention to the AZP (the last added mention to the cluster) produces better results.

$$c_i = \{m_1, m_2, ..., m_z\} \quad (10)$$
$$c_i = \begin{cases} m_1 & \text{the first mention to represent } c_i \\ m_z & \text{the last mention to represent } c_i \end{cases} \quad (11)$$

Next, the AZP and cluster representations are joined together through a concatenation layer. The variable *input* contains the concatenated representation of a mention pair - the AZP and its corresponding cluster. The binary variable *AZP res* receives *input* and is 1 if the AZP and the cluster corefer. The model also outputs the final clusters.

The following equations specify how the output of the network is computed:

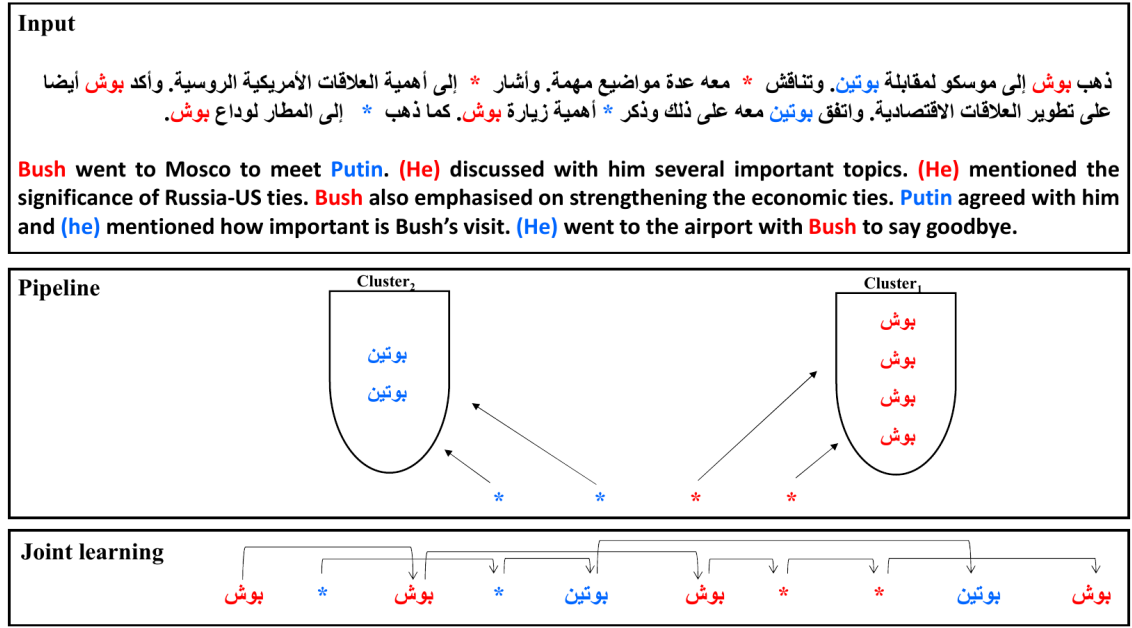

Figure 2: The input is a document and the asterisk * represents the AZPs in the text. For AZP resolution, The *pipeline* resolves AZPs with the output clusters and the *joint learning* resolves AZPs based on coreference chains.

$$input = concat(c_i, a_j) \quad (12)$$
$$input = [c_i, a_j] \quad (13)$$
$$results = AZP\_Res(input) \quad (14)$$
$$results = (r_1, r_2, ..., r_s) \quad (15)$$

The variable *results* consists of the final clusters of the resolved AZPs and non-AZPs. We show the model architecture in Figure 3.

### 4.2 The Joint Learning Model

Our *joint learning* architecture learns to resolve AZPs by using the explicitly represented AZP gaps. This way, AZPs would be learned as any other overt mention. In our extended CoNLL-2012 documents, AZPs have the special identified *pro*. Table 2 shows an example of a CoNLL-2012 original sentence and its extended version. However, we consider AZPs only in the training phase when we apply the coreference resolution model. At test time, AZPs are not considered, same as in a real life application. Instead, we use the *AZP identification* model by (Aloraini and Poesio, 2020b) to tag AZP gaps. After tagging, the input is ready for clustering using the trained coreference resolution model. This is how we represent the inputs for both training and testing:

The *input* is a CoNLL-2012 document with many sentences that has a set of *n* mentions. A mention can be a word or an AZP tag (*pro*).

$$input = (m_1, m_2, m_3, ..., m_n) \quad (16)$$

The variable *input* is fed into the *coreference resolution* (coref_res) model which outputs clusters. The clusters contain mentions and AZPs that refer to the same entity.

$$output\_clusters = coref\_res(input) \quad (17)$$
$$output\_clusters = (c_1, c_2, ..., c_k) \quad (18)$$

For the test phase, we assume a document is not labeled with AZP tags, which reflects real-life applications. Therefore, we first feed *input* into the *AZP identification* (AZP_Id) which outputs *input_with_azp*, that is *input* but with tagged AZPs. The *AZP identification* is pre-trained on the train set of CoNLL-2012 to detect AZP locations.

$$input\_with\_azp = AZP\_Id(input) \quad (19)$$
$$input\_with\_azp = (w_1, a_i, w_2, ..., m_n) \quad (20)$$

After preparing *input_with_azp*, we feed it into the trained *coreference resolution* model which outputs the clusters.

$$results = coref\_res(input\_with\_azp) \quad (21)$$
$$results = (r_1, r_2, ..., r_s) \quad (22)$$

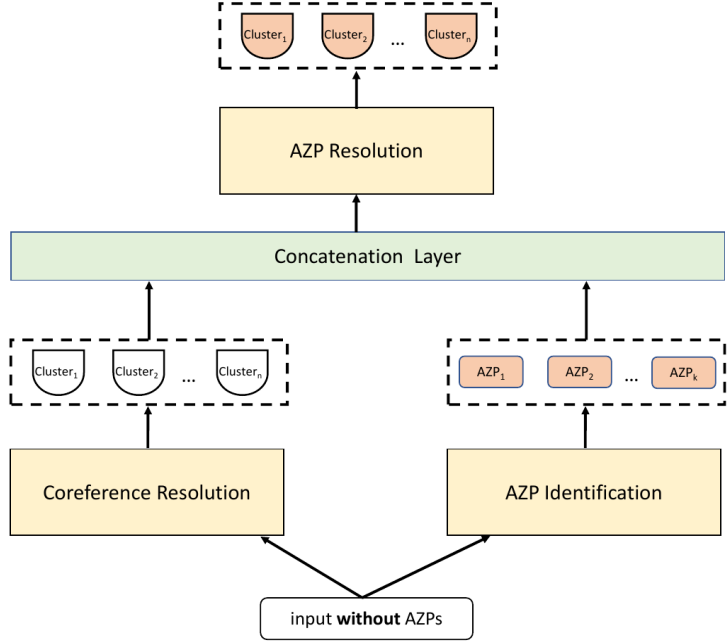

Figure 3: The input without AZPs is fed into the *Coreference Resolution* and *AZP identification* models. The outputs of the two models are clusters and AZPs respectively. Their representations are concatenated, and then their coreference information is learned through the *AZP Resolution* model.

| Original CoNLL-2012 sentence | كانا في الوضع نفسه |
| Extended CoNLL-2012 sentence | كانا *pro* في الوضع نفسه |

Table 2: An example of how we explicitly represent AZPs.

The variable *results* has the resolved AZPs and non-AZPs. We show the overall architecture in Figure 4.

## 5 Evaluation metrics

### 5.1 Coreference resolution

For our evaluation of the coreference system, we use the official CoNLL-2012 evaluation metrics to score the predicted clusters. We report recall, precision, and $F_1$ scores for MUC, $B^3$ and $CEAF_{\phi_4}$ and the average $F_1$ score of those three metrics.

### 5.2 AZP resolution

We evaluate AZP resolution in terms of recall and precision, as defined in (Zhao and Ng, 2007):

$$Recall = \frac{AZPhits}{Number\ of\ AZPs\ in\ Key}$$

$$Precision = \frac{AZPhits}{Number\ of\ AZPs\ in\ Response}$$

*Key* represents the gold set of AZP entities in the dataset, and *Response* represents the predicted resolved AZPs. *AZP hits* are the reported resolved AZP positions in *Response* which occur in the same position as in *Key*.

## 6 Training Objectives

### 6.1 Pipeline

The training objective of the AZP identification is binary cross-entropy loss, as introduced in (Aloraini and Poesio, 2020a):

$$L(\theta) = -\frac{1}{N}\sum_{i=1}^{N}[y_i \log \hat{y}_i + (1-y_i)\log(1-\hat{y}_i)] \quad (23)$$

$\theta$ is the set of learning parameters in the model. $N$ is the number of training samples in the extended CoNLL-2012. $y_i$ is the true label $i$ and $\hat{y}_i$ is its predicted label.

For the AZP resolution, the goal is to minimize the cross entropy error between every AZP and its antecedents, as defined in Aloraini and Poesio's (2020b) model; however, we resolve AZPs with clusters, instead of mentions:

$$L(\theta) = -\sum_{t \in T} \sum_{c \in C} \delta(azp, c) \log(P(azp, c)) \quad (24)$$

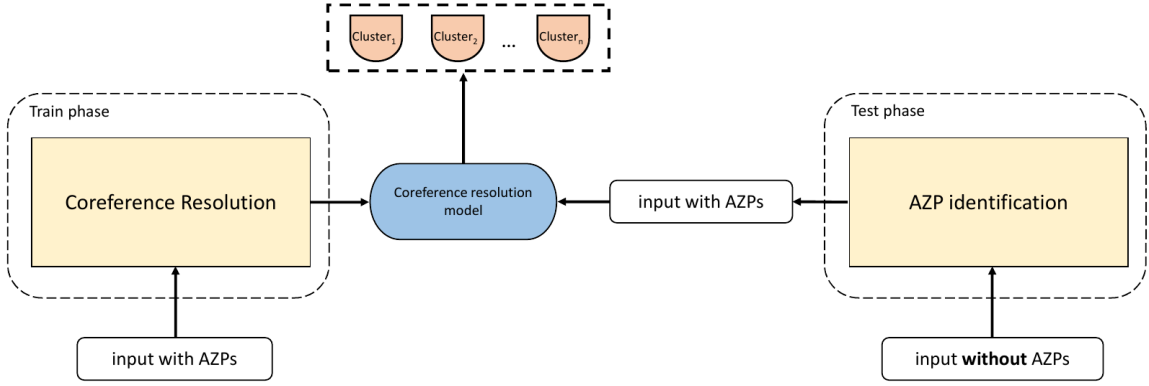

Figure 4: In the train phase, the model learns how to resolve mentions and AZPs. AZPs are represented with the *pro* tag and treated like any other mention. The test phase predicts and tags AZPs locations. We use the model proposed by (Aloraini and Poesio, 2020a) to find AZPs. The pretrained coreference resolution model is used in the test phase to cluster mentions and AZPs.

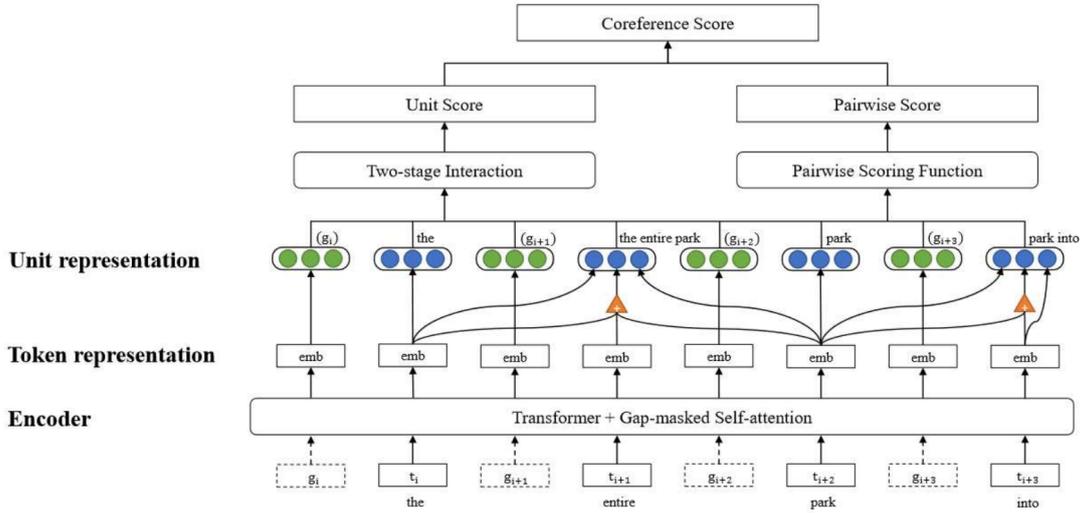

Figure 5: Resolving AZPs and non-AZPs in an end-to-end model (Chen et al., 2021).

*T* consists of the *n* training instances of AZPs, and *C* represents the *k* candidate clusters from the coreference resolution. □*(azp, c)* returns whether a candidate cluster *c* is the correct one for the *azp*, or not. *log(P(azp, c)* is the predicted log probability of the (*azp*, *c*) pair.

The training objective of the coreference resolution is to optimize the log-likelihood of all correct mentions (Lee et al., 2017), as the following :

$$L(\theta) = \log \prod_{i=1}^{N} \sum_{\hat{y} \in \mathcal{Y}(i) \cap G(i)} P(\hat{y}) \quad (25)$$

*G* represents the spans in the gold cluster that includes *i*.

### 6.2 Joint Learning

In the *joint learning*, we only use the (24) for training. AZPs are treated as any other mention; therefore, they become part of the coreference resolution learning objective. We also do not have to train the AZP identification model because we only use the AZP identification in the test phase and we use the pre-trained one on the original CoNLL-2012 from (Aloraini and Poesio, 2020a).

| Models | MUC | | | B³ | | | CEAF$_{\phi_4}$ | | | CoNLL Average |
|---|---|---|---|---|---|---|---|---|---|---|
| | R | P | F$_1$ | R | P | F$_1$ | R | P | F$_1$ | F$_1$ |
| Pipeline | 62.9 | 70.7 | 66.5 | 57.3 | 65.6 | 61.2 | 61.1 | 64.5 | 62.7 | 63.5 |
| Joint learning | 65.2 | 75.5 | 70.0 | 62.6 | 68.3 | 65.3 | 64.8 | 67.7 | 66.2 | 67.1 |
| Chen et al. (2021) | 62.7 | 71.1 | 66.6 | 58.5 | 65.7 | 61.6 | 61.4 | 67.2 | 64.2 | 64.2 |

Table 3: Resolving AZPs and non-AZPs together.

## 7 Results

We compare the results of the *pipeline* and *joint learning* models with the results of Chen et al. (2021). We followed Chen et al. (2021)'s approach for hyperparameter tunning, but we changed the language model to AraBERT-base (Antoun et al., 2020). We evaluate two tasks. First, we assess the results at joint coreference resolution of both AZPs and non-AZPs. Second, we evaluate AZP resolution only. Unlike previous proposals that resolve AZPs with their antecedents, the AZPs of our methods and the Chen et al.'s (2021) model resolve AZPs differently. The *pipeline* uses the output clusters, the *joint learning* uses the coreference chains and Chen et al. (2021) uses two scoring components.

### 7.1 Resolving AZPs and non-AZPs

In Table 3, we see the results of resolving AZPs and non-AZPs. Chen et al.'s (2021) model achieves 64.2% F$_1$ score, which is 0.7% more than the *pipeline*, but less than the *joint learning* with 2.9%. Our *joint learning* approach outperforms our *pipeline* and Chen et al.'s (2021) system, achieving the best F$_1$ average score of 67.1%.

### 7.2 AZP resolution

Next, we compare the AZP resolution results. For the *pipeline*, we used two settings to represent clusters. First, we used the first mention in the cluster to be concatenated with the AZP representation. Second, we used the last-added mention. The *pipeline* approach achieves an F$_1$ score of 58.08% when using the first mention as the cluster representation and 58.59% when using the last mention. The *joint learning* provided better results with an F$_1$ score of 59.33%. Chen et al.'s (2021) model resolved more AZPs correctly than the *pipeline* and *joint learning* methods, achieving an F$_1$ score of 59.49% which is 0.19% more than the *joint learning* score. It seems the two components of Chen et al.'s (2021) model, the Unit Score and Pairwise Score, are able to distinguish AZPs and mentions effectively for the AZP resolution. However, for coreference resolution, they have showed the performance is better when they did not consider AZPs as part of the coreference resolution.

| Training Settings | Test Evaluation | | |
|---|---|---|---|
| | P | R | F$_1$ |
| Pipeline (CR: FM) | 60.34 | 55.98 | 58.08 |
| Pipeline (CR: LM) | 60.97 | 56.39 | 58.59 |
| Joint Learning | 61.41 | 57.40 | 59.33 |
| Chen et al. (2021) | 61.67 | 57.45 | 59.49 |

Table 4: AZP resolution results of *pipeline*, *joint learning* and Chen et al. (2021). FM refers to using the first mention as the cluster representation while LM refers to the last mention.

## 8 Discussion

The main difference between our *joint learning* approach and Chen et al. (2021) is how AZPs are detected and learned. In our approach, we detect AZPs initially before we cluster them with other mentions, while Chen et al.'s (2021) model learns clustering AZPs and mentions in an end-to-end system. Our results appear to confirm earlier results that considering AZP identification end-to-end in the coreference resolution task can negatively affect the performance on the task (Iida and Poesio, 2011; Chen et al., 2021) One possible explanation might be the overall performance of the mention detection on non-AZPs is better than AZPs (Chen et al., 2021). Chen et al. (2021) consider every gap as a candidate AZP, which increases the space of possible candidates and affects their detection recall. To mitigate this problem, we use a different neural component for AZP detection. The AZP identification that we used in the *joint learning* and *pipeline* settings only considers gaps that appear after verbs which limits the number of candidates. Moreover, the AZPs in the *joint learning* have explicit tags which might have resulted in their correct detection, which could be why the approach achieved better results. The main limitation of our proposed approaches is if the AZP identification fails to detect many AZPs in the test phase, it might have dropped the evaluation of the coreference resolution and AZP resolution. Pre-training BERT with AZPs can be beneficial. Existing language models (LMs) learn by masking words or perturbing their order (Qiu et al., 2020), but this is not applicable to AZPs. (Konno et al., 2021) have shown two approaches to improve LMs so they work for AZPs, first by introducing a new pre-training task and second by a new fine-tuning technique. They showed an increased performance for AZP resolution for Japanese. In future works, we intend to pre-train a large-scale LM using their methods and see if it can improve the performance of the AZP and coreference resolution tasks.

## 9 Conclusion

We proposed two architectures to resolve AZPs and non-AZPs jointly. The first approach is in a *pipeline* setting and the second in a *joint learning* representation. The *joint learning* outperformed the *pipeline* and another approach (Chen et al., 2021) in the joint coreference resolution. We also extended the Arabic portion of CoNLL-2012 to include AZPs which will be suitable for future works and shared-tasks that resolves AZPs and non-AZPs together.

## Acknowledgements

We would like to thank the anonymous reviewers for their insightful feedback which helped to improve an early version of the paper.

## A CoNLL-2012 Annotation Layers

The CoNLL-2012 annotation layers consists of the following (Pradhan et al., 2012):

1. Document ID: Contains the file name.

2. Part number: Some files are divided into several files and this number shows the sentence number.

3. Word number: Word position in the sentence.

4. Word itself: This represents the tokenized token.

5. Part-of-Speech: The Part-of-speech of the word.

6. Parse bit: This is the bracketed structure broken before the first open parenthesis in the parse, and the word/part-of-speech leaf is replaced with a *.

7. Lemma: Used to show the gold and predicate lemma.

8. Predicate Frameset ID: This is the PropBank frameset ID of the predicate in Column 7.

9. Word sense: The word sense.

10. Speaker/Author: The speaker or author name, where available. Mostly in broadcast conversation and web log data. However, this is not available for Arabic because all texts are extracted from newspapers.

11. Named Entities: Named entity for the word.

12. Arguments: Predicted and gold arguments.

13. Coreference: Coreference chain which can be single or multiple tokens.